\documentclass[conference,a4paper]{IEEEtran}
\IEEEoverridecommandlockouts
\usepackage{amsmath,amssymb,amsfonts}
\usepackage{algorithmic}
\usepackage{graphicx}
\graphicspath{{img}}
\usepackage{textcomp}
\usepackage{xcolor}
\usepackage{cleveref}
\usepackage{amsmath}
\usepackage{breqn}
\usepackage{url}
\definecolor{darkgreen}{rgb}{0, 0.5, 0.1}

\newcommand{\atrftir}[0]{ATR-$\mu$FTIR}

\begin{document}

\title{Unmixing ATR-µFTIR spectroscopic images of cross-sections of historical oil paintings
\thanks{NN acknowledges the support of the Belgian Federal Science Policy (BELSPO) through the FED-tWIN project Prf-2022-050 BALaTAI.
FMH acknowledges the support of the Belgian Federal Science Policy (BELSPO) through the FED-tWIN project Prf-2021-002 MatCoRe.}
}

\author{
    \IEEEauthorblockN{
        Shivam Pande\IEEEauthorrefmark{1}, 
        Nicolas Nadisic\IEEEauthorrefmark{1}\IEEEauthorrefmark{2}, 
        Francisco Mederos-Henry\IEEEauthorrefmark{2}\IEEEauthorrefmark{3} and 
        Aleksandra  Pi\v{z}urica\IEEEauthorrefmark{1}
    }
    \IEEEauthorblockA{
        \IEEEauthorrefmark{1}Department of Telecommunications and Information Processing, Ghent University, Belgium.
    }
    \IEEEauthorblockA{
        \IEEEauthorrefmark{2}Royal Institute for Cultural Heritage (KIK-IRPA), Belgium.
    }
    \IEEEauthorblockA{
        \IEEEauthorrefmark{3}Engineering of Molecular Nanosystems (EMNS) Laboratory, Université Libre de Bruxelles (ULB), Belgium.
    }

}

\maketitle

\begin{abstract}
Spectroscopic imaging (SI) has become central to heritage science because it enables non-invasive, spatially resolved characterisation of materials in artefacts. In particular, attenuated total reflection Fourier transform infrared microscopy (ATR-$\mu$FTIR) is widely used to analyse painting cross-sections, where a spectrum is recorded at each pixel to form a hyperspectral image (HSI). Interpreting these data is difficult: spectra are often mixtures of several species in heterogeneous, multi-layered and degraded samples, and current practice still relies heavily on manual comparison with reference libraries. This workflow is slow, subjective and hard to scale.

We propose an unsupervised CNN autoencoder for blind unmixing of ATR-$\mu$FTIR HSIs, estimating endmember spectra and their abundance maps while exploiting local spatial structure through patch-based modelling. To reduce sensitivity to atmospheric and acquisition artefacts across more than 1500 bands, we introduce a weighted spectral angle distance (WSAD) loss with automatic band-reliability weights derived from robust measures of spatial flatness, neighbour agreement and spectral roughness. Compared with standard SAD training, WSAD improves interpretability in contamination-prone spectral regions. We demonstrate the method on an ATR-$\mu$FTIR cross-section from the Ghent Altarpiece by the Van Eyck brothers.

\end{abstract}


%%%%%%%%%%%%%%%%%%%%%%%%%%%%%%%%%%%%%%%%%%%%%%%%%%%%%%%%%%%%%%%%%%%%%%%%%%%%%%%%
\section{Introduction}
\label{sec:intro}
\begin{figure}[ht]
    \centering
    \includegraphics[width=0.7\linewidth]{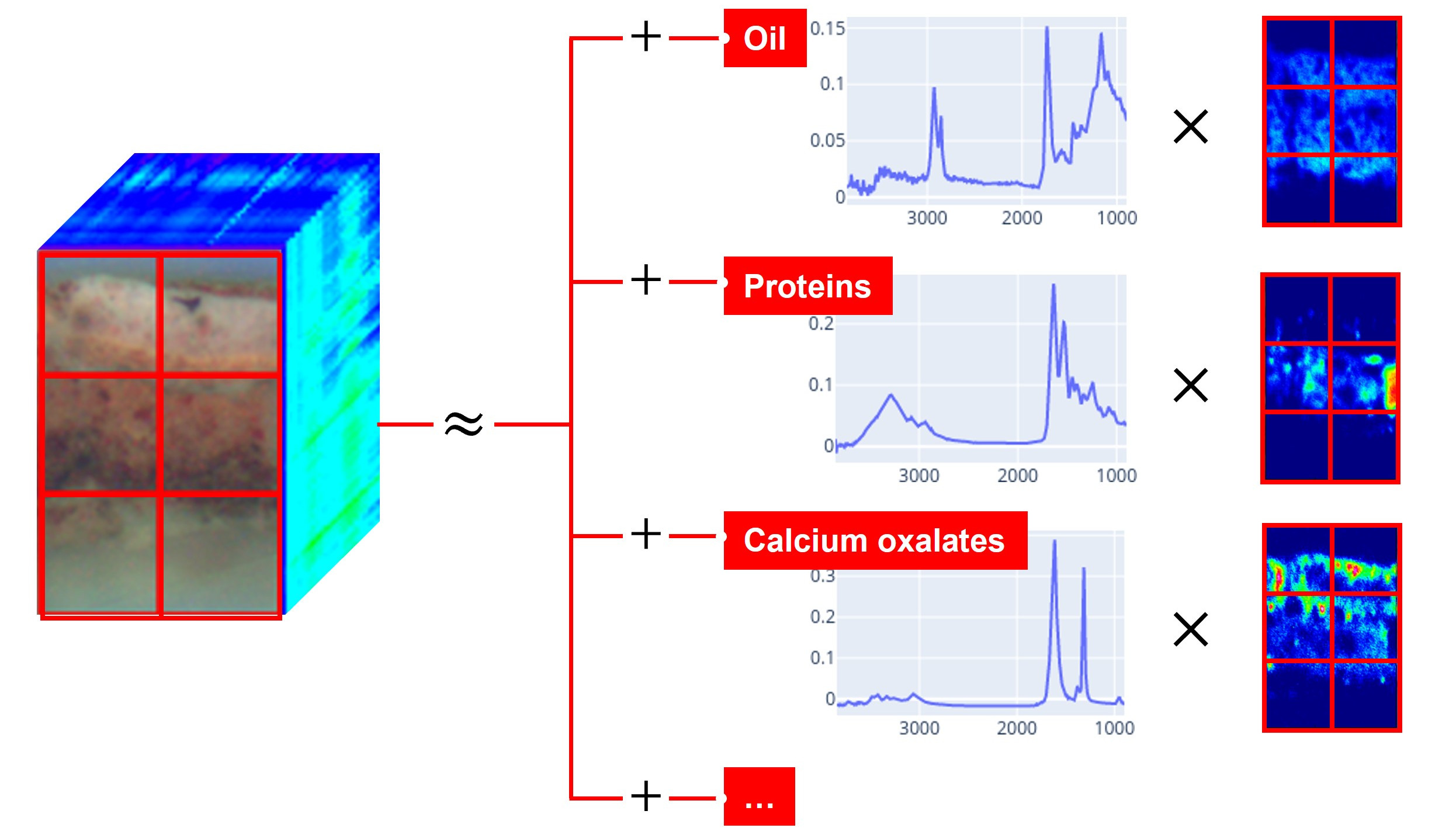}
    \caption{Illustration of an FTIR mapping of a painting's cross-section.}
    \label{fig:muchacho-hsu}
\end{figure}

Over the past decades, spectroscopic techniques have become indispensable tools in the field of heritage science. 
These methods enable researchers to investigate the material composition, manufacturing techniques, and degradation processes of cultural heritage artefacts, offering essential insights for their study, conservation, and restoration~\cite{sober2022revealing, gestels5269502high}. 
A particularly powerful configuration is spectroscopic imaging (SI), which generates spatially resolved chemical maps by acquiring a full spectrum at each pixel across a defined surface \cite{ye2025application}. 
The resulting hyperspectral image (HSI) provides a detailed overview of material distributions within an artefact, revealing compositional patterns that would otherwise remain undetected (Fig.~\ref{fig:muchacho-hsu}).
For all its potential, SI data remains challenging to interpret. 
Cultural heritage artefacts are heterogeneous, multi-layered, and often degraded, so pixel spectra are typically mixtures of multiple species (endmembers). 
In practice, analysts often rely on visual matching to reference libraries, which is time-consuming, dependent on \emph{a priori} expertise, prone to interpretive bias, and frequently insufficient to resolve all components in complex samples. 
These limitations have motivated automated material mapping and identification, and learning-based approaches on hyperspectral reflectance cubes of artworks~\cite{fatmaPigment2025}.

\begin{figure}[ht]
    \centering
    \includegraphics[width=0.85\linewidth]{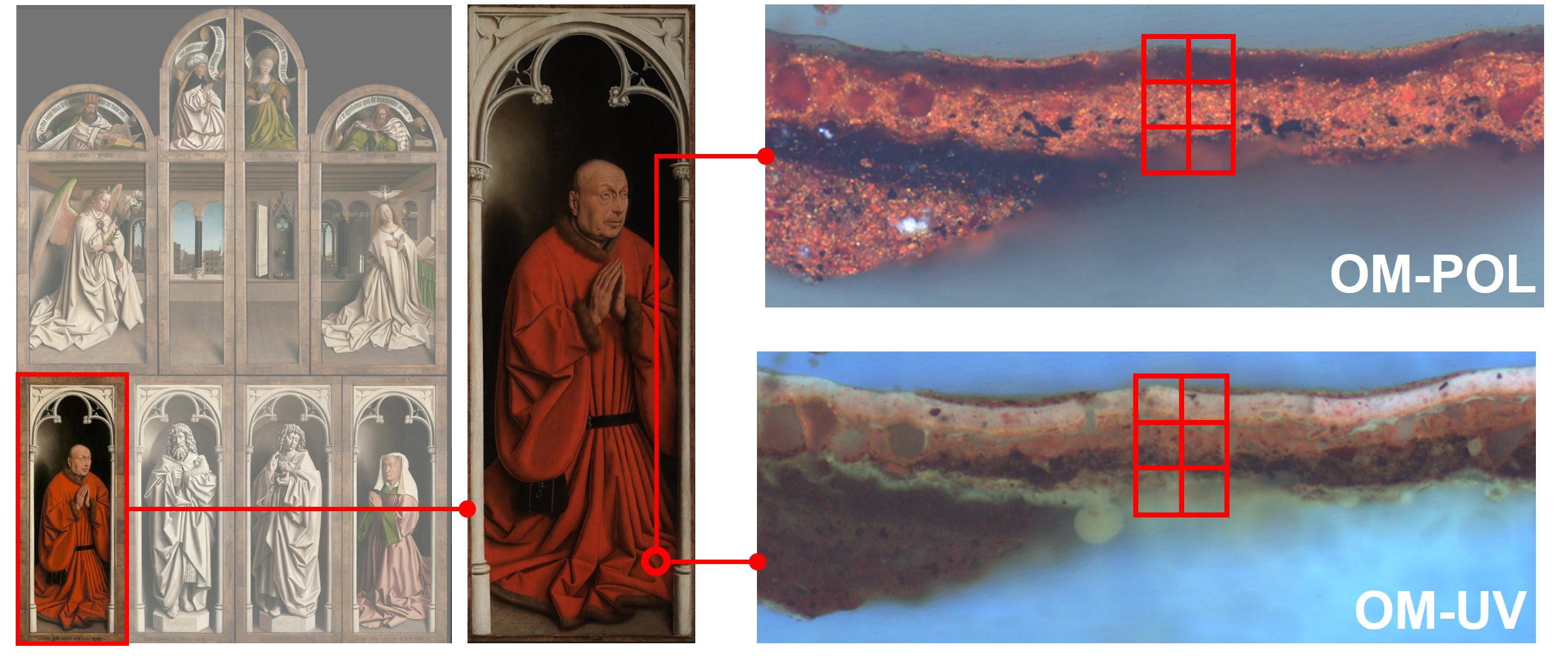}
    \caption{Example of a cross-section observed under polarized (OM-POL) or ultraviolet (OM-UV) light using optical microscopy. The sample was extracted from the point indicated with a red dot in panel XVII, Joos Vijd, belonging to the Ghent Altarpiece \textcopyright KIK-IRPA.}
    \label{fig:muchacho-painting}
\end{figure}

In this work, we focus on attenuated total reflection Fourier transform infrared microspectroscopy (\atrftir{}) \cite{liu2022recent} and its use for the analysis of historical oil paintings' cross-sections. 
Paint cross-sections are a standard preparation in cultural heritage analysis: a millimetric sample containing the stratigraphy is embedded in resin and polished to expose all layers. 
An example observed by optical microscopy under polarized (OM-POL) and ultraviolet (OM-UV) light \cite{mcdonald2019optical} is shown in Fig.~\ref{fig:muchacho-painting}. 
In such samples, dozens of pigments, binders, varnishes and degradation products (e.g., metal soaps or oxalates) may be present, while the analyst often targets only a few, depending on the analytical question.

Hyperspectral unmixing aims to identify endmembers in an HSI and estimate their abundances. 
Traditional approaches rely on the linear mixing model (LMM), where each pixel spectrum is a linear combination of endmember spectra. LMM-based methods include nonnegative matrix factorization (NMF)~\cite{nadisic2023smoothed}, geometric methods such as vertex component analysis (VCA), and sparse regression~\cite{heylen2014review}. 
More recently, deep learning methods have gained traction, leveraging nonlinear representations~\cite{pande2025deep} and spectral--spatial modeling, e.g., CNN-based autoencoders~\cite{palsson2020convolutional}. 
In cultural heritage, deep learning has also been used for pigment unmixing of painting reflectance HSI \cite{rohani2019pigment}. 
However, most prior pigment-HSI works target VIS--NIR reflectance imaging of painted surfaces, whereas \atrftir{} images of cross-sections yields high-dimensional absorbance spectra with different noise, artifacts, and mixing behavior \cite{liu2022recent}.

As a result, existing algorithms are not directly applicable to ATR-$\mu$FTIR cross-section data, which typically exhibits high uncertainty in the number of endmembers, substantial acquisition noise (e.g., due to parasite species such as H$_2$O and CO$_2$), and very high spectral dimensionality (often more than 1500 bands), where not all bands are equally informative. 
Moreover, in reconstruction-based training (as in autoencoders), treating all bands equally can cause noisy, uninformative, or artefact-dominated regions to disproportionately influence the loss and bias the estimated endmembers and abundances. 
To address these challenges, we propose a CNN-based method for blind unmixing of ATR-$\mu$FTIR data. As the main contribution, we introduce a weighted spectral angle distance (WSAD) loss that explicitly accounts for band reliability. 
WSAD downweights unreliable channels using fully automatic band weights derived from (1) spatial flatness (unusually low variance) to downweight uninformative and common-mode bands, (2) local neighbor agreement to penalize isolated channel artefacts, and (3) spectral roughness to suppress spike-like acquisition noise. 
To the best of our knowledge, this is the first proposal for automated unmixing of ATR-$\mu$FTIR data of painting cross-sections.

%%%%%%%%%%%%%%%%%%%%%%%%%%%%%%%%%%%%%%%%%%%%%%%%%%%%%%%%%%%%%%%%%%%%%%%%%%%%%%%%
%========================
\section{Proposed Method}
\label{sec:method}
%========================
For the unmixing problem, we assume a linear mixing model (LMM), since \atrftir{} data adhere to the Beer-Lambert's law of absorbance under realistic assumptions \cite{baij2018time}.
In this section, we first introduce the LMM in a patch-based setting.
Then, we present the proposed method for \atrftir{} unmixing (henceforth called \textit{FTIR-unmixer}).
It builds on a CNN autoencoder~\cite{palsson2020convolutional} and extends to account for the specificities of \atrftir{} data, notably a new loss function called weighted spectral angle distance (WSAD).

\subsection{Linear mixing model (LMM)}
Given a hyperspectral cube with $B$ spectral bands and $K$ endmembers, the standard LMM is
\begin{equation}
\mathbf{x} = \mathbf{E}\mathbf{a} + \boldsymbol{\nu},
\label{eq:lmm}
\end{equation}
where $\mathbf{x}\in\mathbb{R}^{B}$ is the observed spectrum at a pixel, $\mathbf{E}\in\mathbb{R}^{B\times K}$ is the endmember matrix whose $k$-th column is the $k$-th endmember spectrum, $\mathbf{a}\in\mathbb{R}^{K}$ is the abundance vector, and $\boldsymbol{\nu}$ accounts for noise.
In the fully-constrained setting, we enforce the endmember nonnegativity constraint (ENC), abundance nonnegativity constraint (ANC), and abundance sum-to-one constraint (ASC).

\subsection{Patch-wise spectral--spatial LMM}
The pixel-wise LMM in \eqref{eq:lmm} treats pixels independently and ignores spatial structure. To exploit spectral--spatial regularities, we adopt a patch-wise model inspired by \cite{palsson2020convolutional}. Let $\mathbf{X}\in\mathbb{R}^{H\times W\times B}$ denote the hyperspectral data cube, with $H$ and $W$ the spatial dimensions. 
For each spatial location $c$, we consider a $p\times p$ neighborhood (patch) centered at $c$. 
We associate with each pixel in the patch an abundance vector $\mathbf{a}_{c}^{(u,v)}$, resulting in an abundance map $\mathbf{A}_c \in \mathbb{R}^{K\times p \times p}$, where $\mathbf{a}_{c}^{(u,v)} = \mathbf{A}_c(:,u,v)\in\mathbb{R}^{K}$ denotes the abundances at the $(u,v)$-th position within the patch. 
We model the reconstruction of spectra in the patch using a convolutional filters with weights $\mathbf{W}\in\mathbb{R}^{B\times K\times p\times p}$, where each spatial offset $(u,v)$ has its own mixing matrix $\mathbf{W}_{u,v}\in\mathbb{R}^{B\times K}$. The patch reconstruction is written compactly as $\widehat{\mathbf{X}}_c = \mathbf{W} * \mathbf{A}_c$, where $*$ denotes a 2D convolution over $\mathbf{A}_c$. The reconstructed spectrum at the center pixel can be expressed as
\begin{equation}
\widehat{\mathbf{x}}_{c} = \sum_{u=1}^{p}\sum_{v=1}^{p} \mathbf{W}_{u,v}\,\mathbf{a}_{c}^{(u,v)} \;+\; \boldsymbol{\nu}_c,
\label{eq:center_mix}
\end{equation}

\begin{figure*}[ht]
    \centering
    \includegraphics[height=0.31\linewidth]{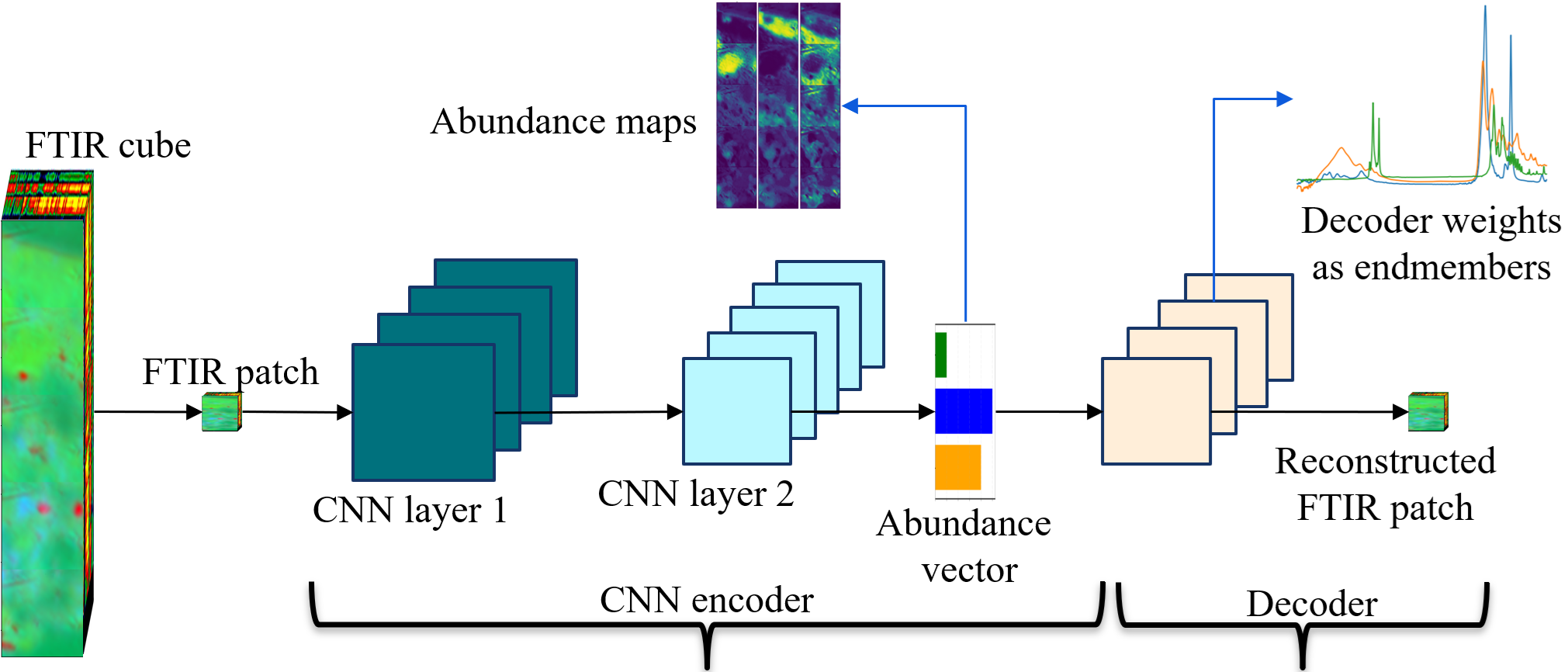}
    \caption{Schematic of FTIR-unmixer. A patch is extracted from the HSI, and sent through the CNN encoder, the outputs of which are interpreted as the abundances. These abundances are sent through a decoder which maps them back to the HSI patch using a linear layer, the weights of which are interpreted as endmembers.}
    \label{fig:schm}
\end{figure*}

\subsection{FTIR-unmixer for patch-wise unmixing}
%We estimate abundances and endmembers using an undercomplete CNN autoencoder composed of an encoder $\mathcal{E}$ and a decoder $\mathcal{D}$. For each input patch $\mathbf{X}_c \in \mathbb{R}^{B\times p\times p}$, the encoder produces abundance maps $\mathbf{A}_c = \mathcal{E}(\mathbf{X}_c)\in\mathbb{R}^{K\times p\times p}$ and the decoder reconstructs the patch as $\widehat{\mathbf{X}}_c = \mathcal{D}(\mathbf{A}_c) = \mathbf{W} * \mathbf{A}_c$. The decoder weights are interpreted as endmember-related quantities through $\mathbf{E}=\sum_{u,v}\mathbf{W}_{u,v}$. The schematic of the proposed method can be visualised in Fig. \ref{fig:schm}.

%\subsubsection{Encoder $\mathcal{E}$ (spectral--spatial abundance estimation)}
%The encoder consists of two convolutional layers:
%\begin{align}
%\mathbf{Z}_1 &= \mathrm{Dropout}\!\left(\sigma\!\left(\mathrm{BN}\!\left(\mathrm{Conv}_{3\times 3}(\mathbf{X}_c)\right)\right)\right), \\
%\mathbf{Z}_2 &= \mathrm{Dropout}\!\left(\sigma\!\left(\mathrm{BN}\!\left(\mathrm{Conv}_{1\times 1}(\mathbf{Z}_1)\right)\right)\right),
%\end{align}
%where $\mathrm{BN}(\cdot)$ denotes batch normalization, $\mathrm{Dropout}(\cdot)$ denotes spatial dropout, and $\sigma(\cdot)$ is a leaky-ReLU activation. The abundance maps are obtained by a scaled softmax to satisfy ANC and ASC: $\mathbf{A}_c(:,u,v) = \mathrm{softmax}\!\left(\alpha\,\mathbf{Z}_2(:,u,v)\right)$, where $\alpha>0$ controls the sharpness of the softmax (larger $\alpha$ encourages sparser abundances).

We estimate abundances and endmembers using an undercomplete CNN autoencoder composed of an encoder $\mathcal{E}$ and a decoder $\mathcal{D}$. 
For each input patch $\mathbf{X}_c \in \mathbb{R}^{B\times p\times p}$, the encoder produces coefficient maps $\mathbf{A}_c = \mathcal{E}(\mathbf{X}_c)\in\mathbb{R}^{K\times p\times p}$, which are interpreted as abundances since each per-pixel vector $\mathbf{A}_c(:,u,v)$ is constrained to lie on the simplex (ANC and ASC) and is used by the decoder as nonnegative mixing coefficients to reconstruct spectra. 
The decoder reconstructs the patch as $\widehat{\mathbf{X}}_c = \mathcal{D}(\mathbf{A}_c) = \mathbf{W} * \mathbf{A}_c$. 
The decoder weights are interpreted as endmember-related quantities through $\mathbf{E}=\sum_{u,v}\mathbf{W}_{u,v}$. 
The schematic of the proposed method can be visualised in Fig. \ref{fig:schm}.

\subsubsection{Encoder $\mathcal{E}$ (spectral--spatial abundance estimation)}
The encoder consists of two convolutional layers:
\begin{align}
\mathbf{Z}_1 &= \mathrm{Dropout}\!\left(\sigma\!\left(\mathrm{BN}\!\left(\mathrm{Conv}_{3\times 3}(\mathbf{X}_c)\right)\right)\right), \\
\mathbf{Z}_2 &= \mathrm{Dropout}\!\left(\sigma\!\left(\mathrm{BN}\!\left(\mathrm{Conv}_{1\times 1}(\mathbf{Z}_1)\right)\right)\right),
\end{align}
where $\mathrm{BN}(\cdot)$ denotes batch normalization, $\mathrm{Dropout}(\cdot)$ denotes spatial dropout, and $\sigma(\cdot)$ is a leaky-ReLU activation. 
The abundance maps are obtained by a scaled softmax to satisfy ANC and ASC: $\mathbf{A}_c(:,u,v) = \mathrm{softmax}\!\left(\alpha\,\mathbf{Z}_2(:,u,v)\right)$, where $\alpha>0$ controls the sharpness of the softmax (larger $\alpha$ encourages sparser abundances).

\subsubsection{Decoder $\mathcal{D}$} %(Positive Convolutional Mixing and Endmember Extraction)}
Decoder is a single linear CNN layer parameterized by an unconstrained tensor $\mathbf{U}\in\mathbb{R}^{B\times K\times p\times p}$. To enforce ENC, we map $\mathbf{U}$ to nonnegative weights using a softplus function: $\mathbf{W} = \mathrm{softplus}(\mathbf{U})$. Reconstruction is performed by $\widehat{\mathbf{X}}_c = \mathbf{W} * \mathbf{A}_c$. Endmember matrix is estimated by summing the spatial slices of the decoder weights:
\begin{equation}
\widehat{\mathbf{E}} = \sum_{u=1}^{p}\sum_{v=1}^{p} \mathbf{W}_{u,v} \in \mathbb{R}^{B\times K}
\label{eq:endmember_sum}
\end{equation}

\subsubsection{Training objective}
The autoencoder parameters are learned by minimizing a reconstruction loss between $\mathbf{X}_c$ and $\widehat{\mathbf{X}}_c$. In this work we use a weighted spectral angle distance (WSAD) to account for band reliability in FTIR data. The full WSAD formulation and the data-driven band-weight estimation are described in the next section.

\subsection{Weighted spectral angle distance (WSAD)}

To reduce the influence of unreliable or noisy spectral regions (e.g., atmospheric absorption), we introduce a nonnegative band-reliability vector $\mathbf{w}\in\mathbb{R}^{B}$ and define WSAD loss by applying element-wise weights before calculating the SAD as shown below:
\begin{equation}
\mathrm{WSAD}(\mathbf{x},\hat{\mathbf{x}};\mathbf{w})=
\arccos\left(
\frac{(\mathbf{w}\odot\mathbf{x})^\top (\mathbf{w}\odot\hat{\mathbf{x}})}
{\|\mathbf{w}\odot\mathbf{x}\|_2\,\|\mathbf{w}\odot\hat{\mathbf{x}}\|_2+\varepsilon}
\right),
\end{equation}
where $\odot$ denotes the Hadamard (elementwise) product. The training loss is the average WSAD over all pixels:
\begin{equation}
\mathcal{L}_{\mathrm{WSAD}} = \frac{1}{N}\sum_{i=1}^{N}\mathrm{WSAD}(\mathbf{x}_i,\hat{\mathbf{x}}_i;\mathbf{w}).
\end{equation}
Here, 
$\mathbf{w}$ is a global band-reliability vector shared across all pixels and kept fixed during training. In Sec.\ref{sec:wt_cal} we compute $\mathbf{w}$ directly from the observed cube using band statistics, to suppress the nuisance regions in the reconstruction objective.

\subsection{Automatic band-reliability weight estimation}
\label{sec:wt_cal}
We estimate $\mathbf{w}$ directly from the input cube using three complementary statistical measures designed to downweight bands that are spatially uninformative or locally spiky.

\paragraph{Flattening and standardization}
Let $\mathbf{Y}\in\mathbb{R}^{N\times B}$ be the flattened cube where row $p$ is $\mathbf{x}_p^\top$.
For each band $b$, define the per-band median and median absolute deviation (MAD) across pixels: $m_b = \mathrm{median}_{p}(Y_{p,b}), \mathrm{MAD}_b = \mathrm{median}_{p}\left(|Y_{p,b}-m_b|\right)+\varepsilon$.
We define a standardized array $\textbf{Z}\in \mathbb{R}^{N \times B}$ with entries as:
\begin{equation}
Z_{p,b}=\frac{Y_{p,b}-m_b}{\mathrm{MAD}_b},
\end{equation}
followed by mean-centering and variance normalization per band to approximate unit variance.

\paragraph{Neighbour-correlation deficit}
Adjacent FTIR bands typically vary smoothly, hence unreliable bands show low adjacent-band correlation ($\rho_b$).
We compute $\rho_b$ as:% (over pixels):
\begin{equation}
\rho_b=\frac{1}{N}\sum_{p=1}^{N} Z_{p,b}\,Z_{p,b+1},\qquad b=1,\dots,B-1.
\end{equation}
For each band, define its local neighbour agreement as $c_1 = \rho_1,\quad c_B=\rho_{B-1},\quad c_b=\min(\rho_{b-1},\rho_b)\;\; \text{for } b=2,\dots,B-1,$ and the neighbour-correlation deficit $d^{(\mathrm{corr})}_b = 1-c_b$. Large $d^{(\mathrm{corr})}_b$ indicates that band $b$ disagrees with at least one neighbour.

\paragraph{Spectral roughness (median spectrum curvature)}
We design a spectral roughness term that targets abnormally high-frequency curvature that is characteristic of band-level artefacts (e.g., single-band spikes, detector glitches, and common-mode atmospheric structures) which can otherwise dominate a reconstruction-driven loss and leak into multiple endmembers. Let $\tilde{\mathbf{m}}\in\mathbb{R}^{B}$ be the median spectrum over pixels: $\tilde{m}_b = \mathrm{median}_{p}(Y_{p,b})$. We quantify per-band roughness using the second finite difference:
\begin{equation}
d^{(\mathrm{rough})}_b = |\tilde{m}_{b+1}-2\tilde{m}_{b}+\tilde{m}_{b-1}|
\end{equation}
where, $2\le b\le B-1$. For $b=1$ $d^{(\mathrm{rough})}_b = d^{(\mathrm{rough})}_2$ and $d^{(\mathrm{rough})}_b = d^{(\mathrm{rough})}_{B-1}$ for $b=B$. Large $d^{(\mathrm{rough})}_b$ indicates unusually sharp curvature or spikiness in the median spectrum. Notably, our proposed roughness term is not intended to suppress genuine chemical peaks, but only the spurious spikes. 

\paragraph{Spatial flatness (low spatial variance)}
Since material composition varies spatially in cross-sections, informative absorptions typically exhibit structured spatial variance, whereas atmospheric or instrument effects can appear as spatially homogeneous modulations. 
We therefore downweight bands with unusually low spatial variance: $v_b = \mathrm{Var}_{p}(Y_{p,b})+\varepsilon$ and $d^{(\mathrm{flat})}_b = -\log(v_b)$. 
Large $d^{(\mathrm{flat})}_b$ corresponds to low variance bands.
%Large $d^{(\mathrm{flat})}_b$ corresponds to low variance (spatially flat/common-mode) bands.

\paragraph{Outlier scoring and weight mapping}
For any vector $\mathbf{u}\in\mathbb{R}^{B}$ we define a score
\begin{equation}
\mathrm{z}(u_b)=\frac{u_b-\mathrm{median}(\mathbf{u})}{\mathrm{MAD}(\mathbf{u})+\varepsilon}.
\end{equation}
We convert the three diagnostics into a nonnegative outlier score to downweight unusually noisy bands
\begin{dmath}
s_b =
\max(\mathrm{z}(d^{(\mathrm{corr})}_b),0)
+
\gamma_{\mathrm{rough}}\max(\mathrm{z}(d^{(\mathrm{rough})}_b),0)
+
\gamma_{\mathrm{flat}}\max(\mathrm{z}(d^{(\mathrm{flat})}_b),0),
\end{dmath}
with hyperparameters $\gamma_{\mathrm{rough}},\gamma_{\mathrm{flat}}\ge 0$. Finally, we map scores to weights using sigmoid around threshold $\tau$:
\begin{equation}
\label{eqn:xw}
w_b = w_{\min} + (1-w_{\min})\;\sigma\!\left(-\alpha(s_b-\tau)\right),
\end{equation}
where $\sigma(t)=\frac{1}{1+e^{-t}}$, $\alpha>0$ controls sharpness, and $w_{\min}\in[0,1)$
sets a minimum weight floor.
Hence, large $s_b$ yields $w_b\approx w_{\min}$, while non-outlier bands yield $w_b\approx 1$.
%%%%%%%%%%%%%%%%%%%%%%%%%%%%%%%%%%%%%%%%%%%%%%%%%%%%%%%%%%%%%%%%%%%%%%%%%%%%%%%%
\section{Experiments and results}
\label{sec:xp}

\begin{figure}[ht]
    \centering
    \includegraphics[height=0.275\linewidth, angle=0]{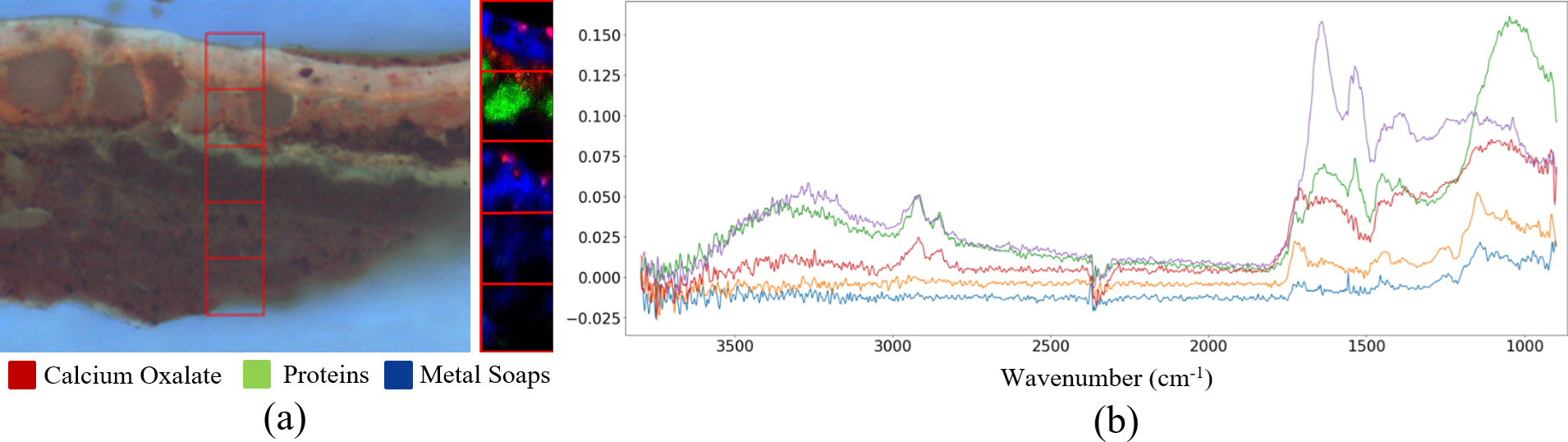}
    \caption{FTIR cross-section of a part from the Ghent altarpiece dataset. (a) We extract $5$ continuous spectral cubes, each of size $64 \times 64$. Three components significant for investigation: calcium oxalates, proteins and metal soaps, along with their possible spatial positions, are presented. (b) The spectra of $5$ randomly selected pixels to show the spectral variation within them.}
    \label{fig:crosssec-ftir}
\end{figure}

In this section, we apply FTIR-unmixer on an \atrftir{} HSI dataset of the cross-section of a sample taken from the Ghent altarpiece (see Fig. \ref{fig:muchacho-painting} and \ref{fig:crosssec-ftir}).
We provide our code through a public repository\footnote{\url{https://github.com/ShivamP1993/ftir-unmixing}}. 

From the painting cross-section, five consecutive tiles are extracted and concatenated along the depth direction.
Each tile is a hyperspectral cube with $64 \times 64$ pixels and $1504$ spectral bands.
Since groundtruth endmember and abundance values are unavailable, we evaluate the two methods qualitatively by spatial and chemical interpretation of the abundances and spectra. 
We focus on spatially localising the calcium oxalates, protein and metal soaps in the paint layers, and comparing it to their relative locations in qualitative reference maps shown in Fig. \ref{fig:crosssec-ftir}a. 
In Fig. \ref{fig:crosssec-ftir}b, we show the spectra of a few random pixels.
In the extracted spectra, we observe that the signal is noisy. Moreover, around 2350 cm$^{-1}$, there are a series of peaks that are not due to the samples composition but rather due to ambient CO$_2$ and H$_2$O present during the measurement. These signals are generally ignored or filtered out by the analysts \cite{liu2022recent}. 
During our experiments, we fix the patch size to $5 \times 5$, while number of extracted patches are set to $20000$. All the models are trained on PyTorch on a $16$GB RTX GPU, using Adam optimizer for $500$ epochs, where the learning rate was empirically fixed to $0.005$. 

\begin{figure}[ht]
    \centering
    \includegraphics[height=0.31\linewidth]{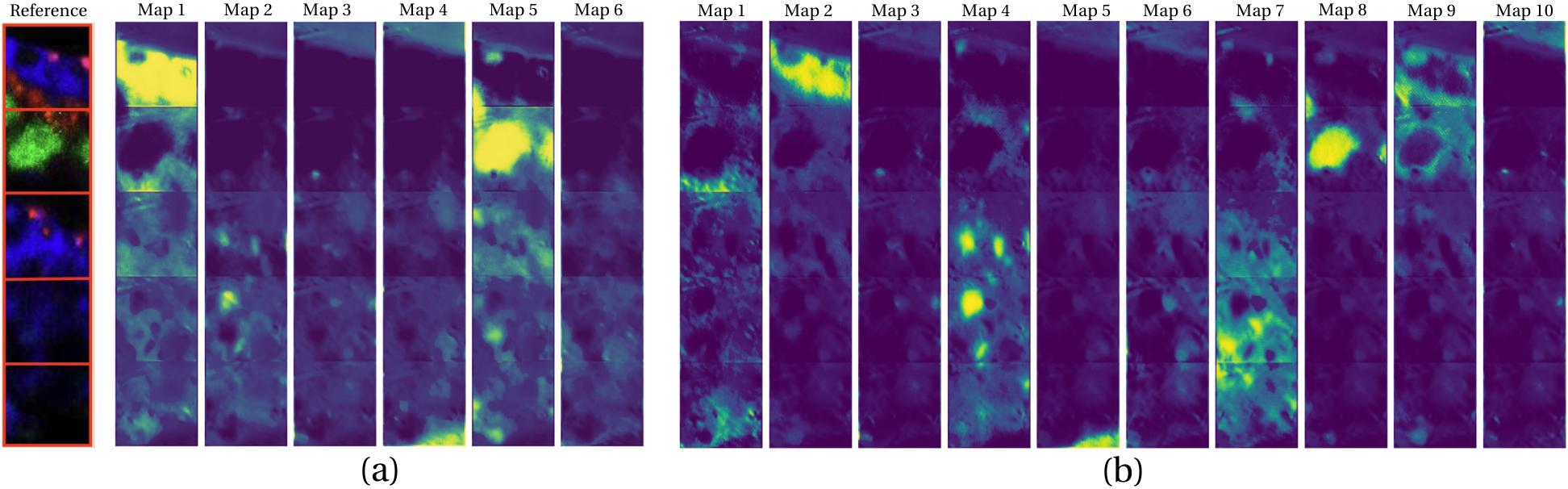}
    \caption{Abundance maps using CNNAE with SAD loss for (a) $K=6$ and (b) $K=10$}
    \label{fig:em_sel}
\end{figure}

The number of endmembers ($K$) is ideally estimated prior to unmixing; however, standard estimators substantially overestimated $K$ for our \atrftir{} cross-section data. 
We selected the endmember count $K$ via an iterative procedure over candidate values for $K\in\{6,\dots,20\}$ and inspected the resulting abundance maps for (i) under-modelling, indicated by merged endmembers appearing within a single abundance map, and (ii) over-modelling, indicated by duplicate endmembers manifesting as repeated, low-intensity maps. 
As $K$ increases, merged maps separate into distinct components (Fig.~\ref{fig:em_sel}a, map $1$, where two endmembers represented in a single map for $K=6$, are differentiated in Fig. \ref{fig:em_sel}b in maps $2$ and $9$, for $K=10$). 
Beyond $K=10$, components begin to duplicate. Based on this criterion, we set $K=10$.

\begin{figure*}[ht]
    \centering
    \includegraphics[height=0.45\linewidth]{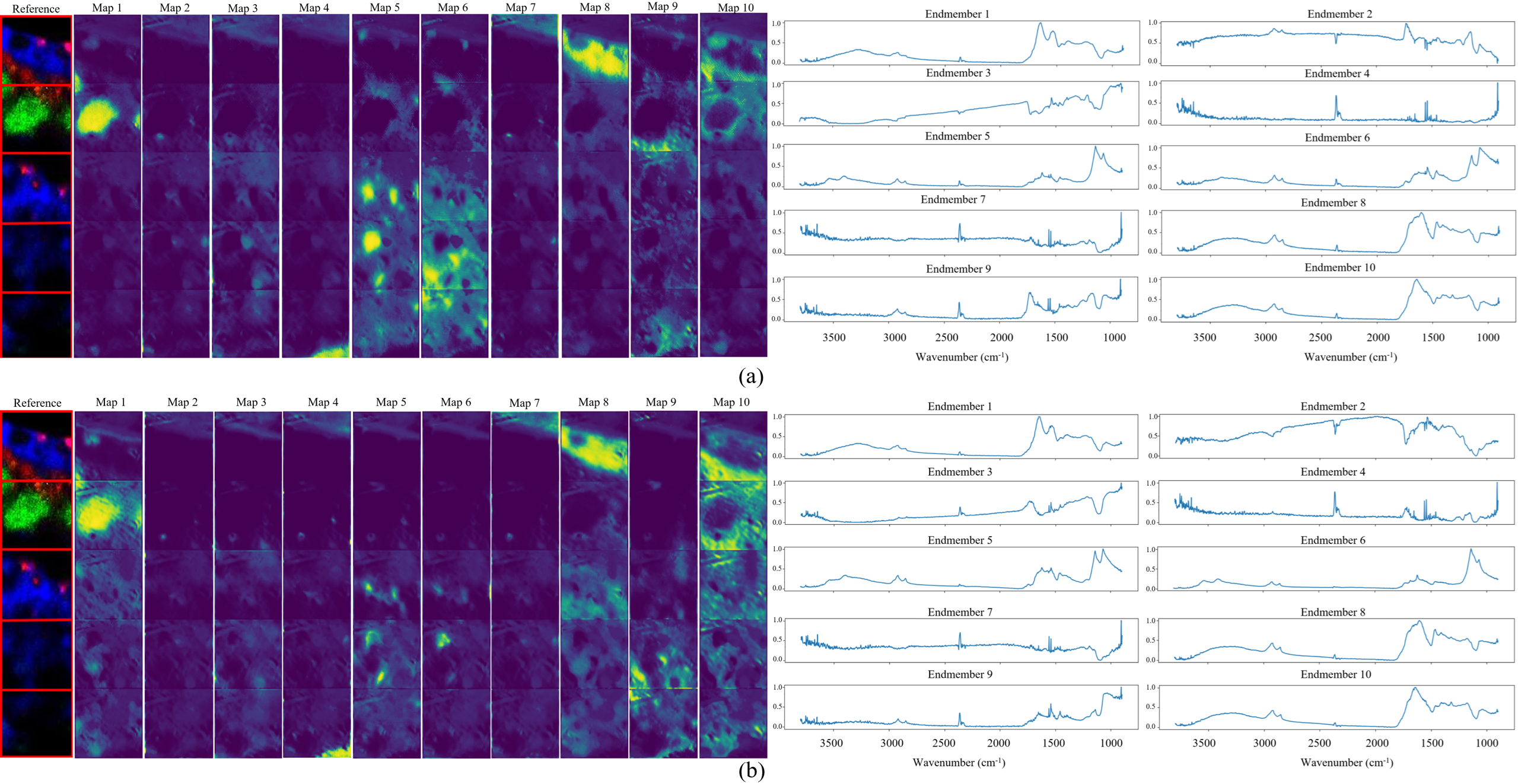}
    \caption{Abundances maps and prospective endmembers from the CNN based method when trained with (a) SAD loss and (b) WSAD loss (ours).}
    \label{fig:maps_and_em}
\end{figure*}

We compare two variants of FTIR-unmixer that share the same architecture and training protocol, differing only in the reconstruction loss: (i) standard SAD, obtained by setting $w_b=1$ in Eq.~\ref{eqn:xw} (equivalent to \cite{palsson2020convolutional}), and (ii) the proposed WSAD with automatically estimated band weights. The resulting abundance maps and learned endmember spectra are shown in Fig.~\ref{fig:maps_and_em}, alongside the reference maps (left column).

In both settings, the method recovers components consistent with key chemical families of interest. In particular, Map~1, Map~8 and Map~10 exhibit spatial patterns consistent with proteins, metal soaps and calcium oxalates, respectively, and the corresponding endmember spectra display characteristic FTIR features for these materials. For the SAD case (Fig.~\ref{fig:maps_and_em}a), proteins (Map~1) and calcium oxalates (Map~10) are reasonably localized, whereas the metal soap distribution (Map~8) appears less uniform across segments. Maps~4 and~7 capture the embedding resin at the top and bottom of the cross-section. 
Maps 5--6 correspond to two previously unspecified components, showing the ability of our method to identify materials without prior knowledge.
Maps~2--3 show a near-zero abundance across the sample, indicating redundant or unused components under the chosen $K$. 
In the endmember spectra, resin-related components (notably Endmembers~4 and~7) exhibit a pronounced feature around $\sim 2350~\mathrm{cm}^{-1}$, attributable to CO$_2$ contamination.

When trained with WSAD (Fig.~\ref{fig:maps_and_em}b), the same main components are recovered (Maps~1, 8 and 10), with a visually improved spatial coherence for metal soaps (Map~8) relative to the reference. Moreover, CO$_2$-related residuals are further suppressed in the endmember spectra associated with the target constituents, consistent with WSAD downweighting unreliable or noisy spectral bands while preserving chemically informative structure. The resin components remain captured in Maps~4 and~7, and the remaining maps again split into secondary materials (Maps~5--6) and unused components (Maps~2--3), while exhibiting reduced sensitivity to the CO$_2$ region.

%%%%%%%%%%%%%%%%%%%%%%%%%%%%%%%%%%%%%%%%%%%%%%%%%%%%%%%%%%%%%%%%%%%%%%%%%%%%%%%%
\section{Conclusion}
\label{sec:conclu}

In this study, we tackled the problem of blind hyperspectral unmixing of \atrftir{} data of paintings' cross-sections.
We introduced a method called FTIR-unmixer, based on a CNN autoencoder and a novel loss function dubbed as WSAD that handles spectral bands of unequal importance as well as noisy bands. It also leverages the spatial smoothness of the abundances.
We studied empirically the performance of our proposal on the unmixing of an \atrftir{} image of a historical oil painting's cross section, and showed that it compares favorably to a state-of-the-art method from the remote sensing community.
To our knowledge, this work is the first to propose an automatic method to unmix \atrftir{} images of cross-sections of historical paintings.

%%%%%%%%%%%%%%%%%%%%%%%%%%%%%%%%%%%%%%%%%%%%%%%%%%%%%%%%%%%%%%%%%%%%%%%%%%%%%%%%
%\section*{Acknowledgment}
\vspace{2cm}
%If needed?
\bibliographystyle{IEEEbib}
{\footnotesize\bibliography{bib2}}

%\printbibliography

\end{document}